\documentclass[runningheads]{llncs}
\usepackage[T1]{fontenc}
\usepackage{graphicx}
\begin{document}
\title{Evaluating LLaMA 3.2 for Software Vulnerability Detection}
%
%
\author{José Gonçalves\inst{1}\orcidID{0009-0004-1038-8384} \and
Miguel Silva\inst{1}\orcidID{0009-0008-6630-9939} \and
Bernardo Cabral\inst{2}\orcidID{0009-0003-1510-7126}\and
Tiago Dias\inst{1}\orcidID{0000-0002-1693-7872} \and
Eva Maia\inst{1}\orcidID{0000-0002-8075-531X}\and
Isabel Praça\inst{1}\orcidID{0000-0002-2519-9859} \and
Ricardo Severino\inst{2}\orcidID{0000-0002-4215-3238} \and
Luís Lino Ferreira\inst{2}\orcidID{0000-0002-5976-8853}
}
%

\authorrunning{J. Gonçalves et al.}
%
\institute{Research Group on Intelligent Engineering and Computing for Advanced Innovation and Development (GECAD), School of Engineering of the Polytechnic of Porto, 4249-015, Porto, Portugal \email{\{jpsgs,mdgsa,tiada,egm,icp\}@isep.ipp.pt} \and
INESC TEC/Polytechnic of Porto - School of Engineering, Porto, Portugal
\email{\{bemac,sev,llf\}@isep.ipp.pt}\\
}
\maketitle              
\begin{abstract}
Deep Learning (DL) has emerged as a powerful tool for vulnerability detection, often outperforming traditional solutions. However, developing effective DL models requires large amounts of real-world data, which can be difficult to obtain in sufficient quantities. To address this challenge, DiverseVul dataset has been curated as the largest dataset of vulnerable and non-vulnerable C/C++ functions extracted exclusively from real-world projects. Its goal is to provide high-quality, large-scale samples for training DL models. However, during our study several inconsistencies were identified in the raw dataset while applying pre-processing techniques, highlighting the need for a refined version. In this work, we present a refined version of DiverseVul dataset, which is used to fine-tune a large language model, LLaMA 3.2,  for vulnerability detection. Experimental results show that the use of pre-processing techniques led to an improvement in performance, with the model achieving an F1-Score of 66\%, a competitive result when compared to our baseline, which achieved a 47\% F1-Score in software vulnerability detection.

\end{abstract}
\section{Introduction}\label{sec:intro}

As software adoption grows, so do concerns about its security. In modern software development, these risks are worsened by the use of AI-powered code assistants that can generate insecure code that can be deployed in production environments \cite{Toth2024,Perry2023}. Thus, it is of paramount importance to ensure the quality of software, largely due to the fact that  it serves as the backbone of many critical operations \cite{Malhotra2020,Shen2020,Ghaffarian2017}.

One of the key approaches to maintain software quality is software testing, which provides insight into the development process to assess whether the software is well developed \cite{Maltaie2020}. This field encompasses a wide range of methods, each designed to evaluate different quality attributes \cite{Ganney2020}. Among these methods, Software Vulnerability Detection (SVD) plays a critical role in identifying security vulnerabilities. Static analysis, a common technique for SVD, examines source code for vulnerabilities without executing it, relying on methods such as rule/template-based analysis, static symbolic execution, and code similarity detection \cite{Malhotra2020,Shen2020,Ghaffarian2017,Shafiq2021,Zhu2022}. In order to reduce the time and cost associated with manual vulnerability analysis, extensive research has been conducted on automating SVD using Deep Learning (DL). Studies have explored different models and feature representations using structural, syntactic, and semantic information extracted from source code \cite{Zhu2022,Wang2021}. Despite achieving high performance, DL-based SVD still faces several challenges, mainly due to the complexity of source code and the scarcity of high-quality labeled datasets \cite{Shen2020,Akimova2021}.

To address the complexity of source code, several studies have explored multiple feature representations at different levels of granularity. Most modern approaches reduce data complexity by pre-processing and extracting implicit structural, syntactic and semantic features from the source code \cite{Akimova2021}. Successful methods remove non-essential elements such as comments and variable names \cite{Zou2021}, while normalising code components that contribute to vulnerabilities, thus simplifying the samples. The presence of programmer-defined elements can also affect model performance. Experimental results \cite{ALERT_2022} indicate that Artificial Intelligence (AI) models often rely on identifier names rather than the underlying code semantics associated with vulnerabilities. This reliance introduces a security risk, as malicious actors can create adversarial samples to manipulate the model's output by changing identifier names while preserving the actual vulnerability. Graph representations have also been explored, specifically Abstract Syntax Trees (ASTs), Control Flow Graphs, and Code Property Graphs, to effectively represent the semantic and syntactic information hidden in source code~\cite{Zhu2022,Wang2021,SUN2024103732}.

In recent years, many research works have attempted to fine-tune Large Language Models (LLMs) on source code for SVD, as will be described in Section~\ref{sec:sota}. However, the most promising results are achieved using synthetic datasets, which do not capture the full complexity of real-world source code~\cite{Jain2023}. As a result, these models often underperform when applied to real-world source code. Driven by the lack of valuable real-world data, a dataset named DiverseVul~\cite{Chen2023}, containing vulnerable and non-vulnerable C/C++ source code collected from several real-world vulnerable projects, was proposed. The authors of the dataset analyzed it in combination with other well-known ones, and even though they present discouraging results, they mention that the combination of the datasets could have turned some data points miscellaneous, which would affect performance~\cite{Chen2023}.

This work aims to study the DiverseVul dataset in the context of LLMs for SVD, while also exploring the influence of pre-processing techniques. LLaMA 3.2, a recent model released by Meta, was fine-tuned for SVD using the DiverseVul dataset. This text-only lightweight model allows for the development of solutions that can be executed in lower end devices, while achieving good results. The performance of this model for SVD was analyzed both before and after the application of source code pre-processing techniques, to assess if the model is influenced by non-relevant code elements. This pre-processing was conducted using the Source Code Processing Engine (SCoPE)~\cite{SCoPE2024}, a C/C++ data processing framework that operates at the function level. During the data processing phase, several erroneous data points in DiverseVul were identified and removed, resulting in the creation of a refined version of the original dataset.

In summary, the main contributions of this work include:

\begin{itemize}
    \item a refined version of the DiverseVul dataset processed by the SCoPE framework without the erroneous entries. The refined version of the dataset is publicly available on GitHub.
    \item an assessment of the influence of pre-processing on vulnerability detection with LLMs.
    \item a verification of whether newer and lighter models, such as LLaMA 3.2 with 1 billion parameters, can achieve good performance in SVD.
\end{itemize}

This work is organized as follows. Section~\ref{sec:sota} provides a literature review of existing LLMs applications for SVD. Section~\ref{sec:exp_setup} presents the experimental setup, describing the dataset and its flaws, the data pre-processing and the LLaMA 3.2 model. Section~\ref{sec:results} presents and discusses the results achieved before and after the data processing phase. Section~\ref{sec:threats} addresses the potencial threats to validity. Finally, Section~\ref{sec:conclusions} summarises the work with the main conclusions drawn from it and describes new research lines to be explored in future work.

\section{State of the Art}
\label{sec:sota}

SVD is the process of identifying weaknesses or flaws within software systems. It can be performed via two different analysis methods: (i) static and (ii) dynamic\cite{Zhu2022}. Static analysis consists of analyzing the source code to detect vulnerabilities. This method has high code coverage but is usually more prone to false positives \cite{Zhu2022}. Dynamic analysis is performed by running a set of test cases in the target system and observing its behavior in real-time to ensure compliance with specified requirements\cite{Zhu2022}. This method has higher miss rate and achieves less coverage, but it is generally more accurate and raises fewer false alarms\cite{Zhu2022}. In comparison to one another, dynamic analysis requires a substantial amount of computational power to generate and run multiple test cases in the System Under Test (SUT), whereas static analysis does not. However, static analysis requires a great amount of knowledge and attention, which can be very laborious. To circumvent this process and decrease the amount of expertise necessary to perform this static vulnerability detection, several research works have attempted to leverage DL to automate the vulnerability detection process \cite{Li2019,TANG2024103816}.

The literature shows that DL has been widely adopted to perform static analysis of source code, experimenting several feature representations for ingestion and algorithms, ranging from the most simple Neural Networks (NN) to more complex model architectures, such as Transformers in the case of LLMs. For instance, Russel et al. \cite{Russel2018} used multiple types of NN to detect software vulnerabilities from C/C++ source code extracted from Debian, Github, and SATE IV Juliet Test Suite repositories represented by token embeddings. The Convolution Neural Network (CNN) had the best performance on the SATE IV dataset, and the CNN combined with the random forest showed the most promising results on the Debian and Github datasets. Their work proved that machine learning is more effective than the use of static analyzers, which are typically rule-based. Similarly, Bilgin et al. \cite{Bilgin2020} decided to experiment with the CNN, but instead of using the token representation, the authors chose to leverage ASTs to represent the source code. The AST ensured a structured format suitable for the model, with the source code transformed into a binary tree of uniform depth for consistent function size. The binary tree was then converted into a depth-ordered token vector and further transformed into a three-dimensional numeric vector using token embeddings. The study, conducted on a dataset with predetermined vulnerabilities, yielded promising results, noting variations in the difficulty of detecting different vulnerability types. 

Other works continued to explore more data representation of source code that would contain more information than the one provided by ASTs. Li et al. \cite{Li2018} explored the impact of incorporating semantic code information in detecting vulnerabilities associated with specific library/API calls in source code. They compared the effectiveness of classification using only data dependency features with the simultaneous use of control dependency and data dependency features. The Joern tool was employed to analyze programs, extract features, identify library/API function calls, and generate code gadgets from program slices. These gadgets were vectorized and tested with different NNs. The study found that combining control and data dependency features is more effective than relying solely on data dependency. The best performance was achieved by a Bidirectional Long Short-Term Memory (BLSTM) NN, with an F1-Score of 92.4\% on a dataset of 68353 code gadgets, including 13686 labelled as vulnerable. The authors tested various methods to handle imbalanced data and found that not using any sampling produced the best results.

Zou et al. \cite{Zou2021} proposed µVulDeePecker, a deep learning-based system for multi-classification of software vulnerabilities. Inspired by VulDeePecker \cite{Zhen2018}, the first binary vulnerability detection system based on deep learning, µVulDeePecker refines the code gadget concept. Unlike VulDeePecker, which proved ineffective for multiclass vulnerability detection, µVulDeePecker introduces two variants. The first, VulDeePecker+, modifies the NN architecture directly. The second trains separate VulDeePecker classifiers for each vulnerability type. However, the latter variant faces scalability issues and is less effective with small sample sizes. µVulDeePecker comprises six modules, including data pre-processing for cleaning and normalizing input data, classification of collected data into vulnerability classes, and an automatic program based on lexical analysis to analyze tokens. Vectors representing normalized code gadgets are generated using a word-to-vector model. The BLSTM NN architecture, built using Keras, is trained separately for global and local feature learning to prevent interference between networks. Using the same dataset as Li et al. \cite{Zhen2018}, µVulDeePecker achieved a maximum F1-Score of 96.28\%, demonstrating its effectiveness and improvement over its predecessor.

Zhou et al. \cite{Zhou2019} introduced Devign, a model utilizing graph structures to learn vulnerability patterns from source code functions through Graph Neural Networks (GNNs). To mitigate dataset inaccuracies resulting from static tool classifications, they produced their own dataset containing labelled C functions from open-source projects like Linux Kernel, QEMU, Wireshark, and FFmpeg. The Devign architecture includes a graph layer representing the semantics of composite code, a Gated Graph Recurrent layer learning node characteristics from neighboring nodes, and a Conv module extracting significant node representations for graph-level prediction. The results show that Devign, with comprehensive graph-encoded semantics, outperforms most modern vulnerability identification methods. It achieved a maximum F1-Score of 84.97\% on Linux function data following best practices. For the FFmpeg dataset, Devign obtained a lower maximum F1-Score of 34.92\%, while on the Qemu dataset, it achieved a higher F1-Score of 41.12\%.

With the advances in AI, LLMs have also gained some popularity in the vulnerability detection field. For instance, Khare et al. \cite{Khare2023} evaluate the performance of LLMs from the GPT and LLaMA families on 5 different datasets spanning two programming languages and including code samples from synthetic and real-world projects. The authors obtained great results on the synthetic datasets, with smaller fine-tuned LLMs performing better than larger models and lower results on the real-world projects datasets which include CVEFixes Java and C/C++ subsets. Nevertheless, their results show that LLMs can outperform existing static analysis and deep learning vulnerability detection tools by identifying the vulnerable data flows in the code. Chen et al. \cite{Chen2023}, responsible for the DiverseVul dataset, also combined multiple datasets including their own, and fine-tuned LLMs of four different families to assess their quality. The results achieved by the authors show that NatGen, CodeGPT, CodeBERT, and ReVeal were the best-performing algorithms of each family, with NatGen achieving the highest F1-Score of 47.2\%. However, their approach is prone to errors since the combination of datasets tailored differently might result in miscellaneous data. Additionally, the authors do not seem to perform any data balancing, which could explain the high accuracy and low F1-Score. Mamede, Pinconschi, and Abreu \cite{Mamede2023} fine-tuned JavaBERT from the BERT model architecture family to perform multi-label classification. The authors use the Juliet Test Suite for Java, composed of synthetic data covering 21 different vulnerability types, to fine-tune the model and its evaluation presents an accuracy of 98.9\%. 

Though the results seem promising, multiple works have shown that synthetic datasets in the field do not represent the code implemented in real projects \cite{Alexopoulos2022,Lipp2022,Katsadouros2022,Steenhoek2022}. Driven by the lack of valuable data, Kuang et al. \cite{Kuang2023} mutated user-defined identifiers in the source code while maintaining its syntactic and semantic information to generate counterfactual training data. They implement their data processing method on VulDeePecker, Draper, and CodeXGLUE datasets and evaluate them on CodeBERT. Their results show that their method performed better on Draper and VulDeePecker datasets, achieving 94\% and 85\% F1-Score, respectively. However, these are both semi-synthetic datasets. 

Alternatively, some works have also leveraged prompt engineering techniques to perform vulnerability detection. For instance, Steenhoek et al. \cite{Steenhoek2022} evaluated the performance of 11 well-known LLMs and compared their performance on vulnerability detection on a dataset curated from real-world data collected from Common Vulnerabilities and Exposures (CVE) records. The results show that the LLMs achieved 50\% to 63\% F1-Score. Purba et al. \cite{Purba2023} decided to use CVEFixes for fine-tuning LLMs for vulnerability detection on two types of attacks, SQL injection and buffer overflow. However, their results using CodeGen, DaVinci and GPT-4 and GPT3.5 show low performance, similar to other literature works. Zhang et al. \cite{zhang2024} planned to explore further prompt engineering for SVD using specifically ChatGPT and proceeded to complement previous works in the field by improving the design of the prompts utilized. They experimented on two datasets containing Java and C/C++ source code. In regards to vulnerability detection in the C/C++ code, their chain-of-thought prompts improved performance to 74\% F1-Score.

\section{Experimental Setup}\label{sec:exp_setup}

The study conducted in this work explores the DiverseVul dataset in the field of SVD by fine-tuning a pre-trainned model for vulnerability detection, while also analysing the impact of the data pre-processing. The experiments were conducted on a laptop running Windows 11 OS, equipped with a 12th Gen I7-12650HX 2.1 GHz CPU, 64GB of RAM, and an RTX A4500 GPU. The following subsections describe the most relevant aspects of the experiment's setup.

\subsection{Dataset and Data Pre-processing}

DiverseVul is a comprehensive vulnerability dataset that includes 18945 vulnerable C/C++ across 150 Common Weakness Enumerations (CWEs). Complementing these are 330492 non-vulnerable functions sourced from 7514 vulnerability-fixing commits. The dataset's curation involved the crawling of two security issue websites and the extraction of data from the vulnerability fix commits of 797 vulnerable projects. Commits implying vulnerability introduction were excluded using regex and manual verification for commits with more than 10 functions altered. Projects related to these commits were cloned, and their vulnerable functions were extracted and labelled as vulnerable or non-vulnerable before and after the commits, respectively. Duplicate entries were removed via MD5 hash comparison. Functions lacking specific CWE labels were cross-referenced with CVE numbers from the titles, aligning them with respective CWEs in the National Vulnerability Database (NVD). The final dataset is composed of 8 features that are described in Table~\ref{tab:dataset_features}.

\begin{table}[ht]
\centering
\caption{DiverseVul Dataset Features}\label{tab:dataset_features}%
\begin{tabular}{@{}ll@{}}
\hline
\textbf{Feature} & \textbf{Description} \\
\hline
func & Contains the C/C++ function code. \\
target & Does the function is vulnerable or not.  \\
cwe & List of respective CWEs present in the function. \\
project & Project from where the function was extracted. \\
commit\_id & Identifier of the commit. \\
size & Size of the function. \\
message & Commit message. \\
\hline
\end{tabular}
\end{table}

AI models may rely on programmer-controlled elements to predict whether a function is vulnerable, potentially overlooking the actual vulnerability. To address this, the SCoPE framework was used to process all source code from DiverseVul dataset. SCoPE facilitates code standardization by leveraging regular expressions. It efficiently removes redundant white spaces and comments while substituting specific strings with generic tokens. 
It also uses token substitution to streamline both function and variable declarations, significantly reducing code size. This ensures a consistent representation of the data, free from superfluous elements that could impact the model's performance. To apply the transformations, the code is first converted into a representative tree, which is used to detect variables and function names. Then, token replacement and code normalization are performed using regular expressions, resulting in a normalized representation of the source code. 

\begin{figure}[htbp]
\centering
\includegraphics[width=\columnwidth,bb=0 0 932 193]{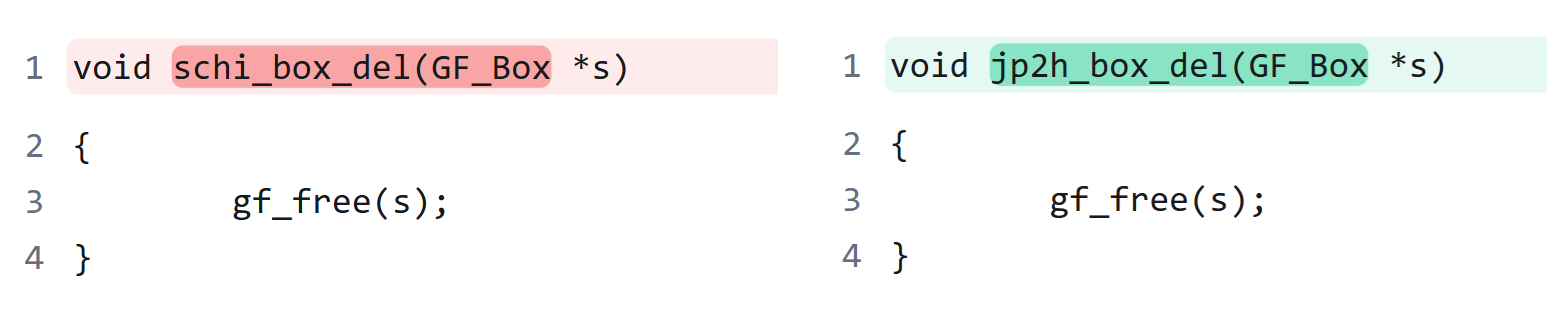}
\caption{Example of Duplicated Entries}
\label{fig:duplicated_entries_example}
\end{figure}

Following the application of SCoPE to Diversevul, 268253 entries were successfully processed without problems, while 62239 entries were processed with errors due to the use of macros in the code.
Subsequently, an additional step was taken to identify functions sharing identical declarations. This precautionary measure aimed to validate the integrity of the data curation process.  As a result,  7901 duplicated entries were identified, and faulty data was detected. Examples of faulty data included entries containing only comments or entries that, despite having identical code, differed only in function names while having opposite labels. One of the faulty entries can be observed in Fig.~\ref{fig:duplicated_entries_example}, where two entries have the same code with different function names. 

After identifying and removing erroneous entries, all entries containing only comments were eliminated. Duplicate entries were also identified and removed to ensure that there was only one unique example for each entry. After this pre-processing, a new refined version of the DiverseVul dataset, now free of such entries, was created and is publicly available on GitHub for further use and exploration. 

\begin{figure}[htbp]
\centering
\includegraphics[width=0.8\columnwidth, bb=0 0 519 380]{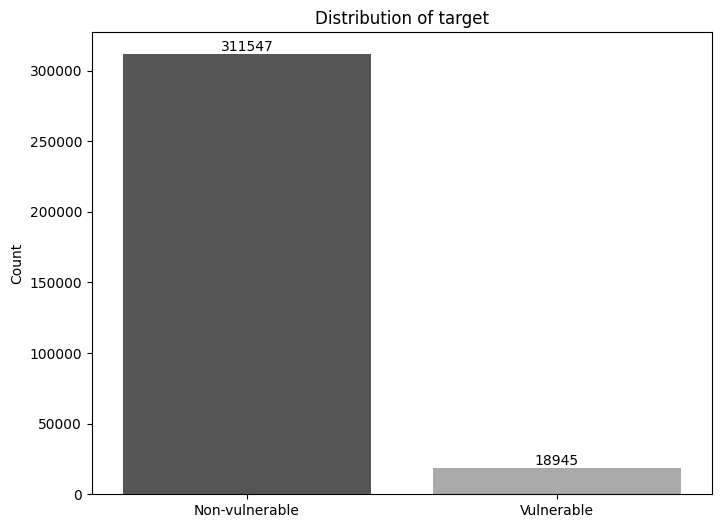}

\caption{DiverseVul Target Distribution}
\label{fig:data_distr}
\end{figure}

Given the significant class imbalance in the dataset, as shown in Fig.~\ref{fig:data_distr}, a balanced subset of 10000 entries was selected for training and evaluation of our model in the SVD task, consisting of 5000 samples from each class. This selection aimed to optimize computational efficiency while maintaining the representativeness of the dataset. To evaluate model performance, the holdout method was used to randomly partition the data into training, test and validation sets. The training set contained 70\% of the samples, while the remaining 30\% were split between the validation and test sets. 

\subsection{Model and Fine-tuning}

This study aims to assess if LLaMA 3.2 with 1.23 billion parameters is effective at SVD, both before and after optimizing the input data. Despite the original purpose of the LLaMA 3.2 model was sequence-to-sequence generation, originally pre-trained on a large multilingual dataset for general language tasks, in this study, the model was fine-tuned to perform binary classification on code samples, categorizing them as either vulnerable or not vulnerable. Due to computational constraints, low-rank adaptation (LoRA) was used in the fine-tuning process. LoRA is a Parameter-Efficient Fine-Tuning (PEFT) technique that reduces the number of trainable parameters, significantly reducing training time while maintaining model performance \cite{Razuvayevskaya2023Comparison}.


To optimize the model for vulnerability detection, multiple fine-tuning iterations were performed to identify the best hyperparameter configuration. The final set of hyperparameters that achieved the best performance is detailed in Table~\ref{tab:hyperparameters}.

\begin{table}[ht]
    \centering
    \caption{Model Hyperparameters}\label{tab:hyperparameters}
    \begin{tabular}{@{}ll@{}}
    \hline
    \textbf{Hyperparameter}       & \textbf{Value} \\
    \hline 
    Epochs                        & 4                         \\
    Batch Size                    & 32                        \\
    Learning Rate                 & \(2 \times 10^{-5}\)      \\
    Weight Decay                  & 0.01                      \\
    Evaluation Interval           & Every 100 steps           \\
    \hline
    \end{tabular}
    \end{table}

\section{Results and Discussion}
\label{sec:results}


One of the main challenges faced by some LLMs in SVD is their limited token capacity, which limits the amount of code they can process at a time \cite{zhang2024extendingllmscontextwindow}. When the length of the analyzed source code exceeds this limit, the model's ability to capture interrelationships between different code segments may be compromised \cite{li2024longcontextllmsstrugglelong}. To address this issue, reducing the number of tokens in the input not only mitigates this limitation, but also improves the model's performance, allowing it to more effectively detect vulnerabilities in larger segments of code. This approach streamlines both the training and inference processes \cite{song2024hierarchicalcontextmergingbetter}.

In this context, since the dataset had an average token length of 374.13 tokens per function, SCoPE was used to reduce the token length, resulting in 300.38 tokens per function after processing. Although the model used in the experiments supports significantly larger context windows (up to 128,000 tokens), a limit of 1024 tokens was set in the tokenizer for the maximum size allowed per function. This limit was set to reduce the need for padding in smaller functions, optimizing the use of computational resources during training. Only 5.48\% of the functions in the original dataset exceeded this limit, and this proportion was further reduced with SCoPE processing. Reducing the average length of the functions minimized the cases of truncation in the tokenization process.


The model was initially trained on just two epochs to quickly assess whether SCoPE would have a positive impact without the computational cost of full fine-tuning. Due to the long training time, full tuning of all setups was impractical, so only the best performing setup was fine-tuned to refine its performance.

\begin{table}[!htbp]
    \centering
    \caption{Obtained results for DiverseVul.}
    \label{tab:results_proc}
    \begin{tabular}{l@{\hskip 0.5cm}l@{\hskip 0.5cm}l@{\hskip 0.5cm}l@{\hskip 0.5cm}l@{\hskip 0.5cm}}
    \hline
    \textbf{Processed} &\textbf{Accuracy} & \textbf{Precision} & \textbf{Recall} & \textbf{F1-Score} \\ \hline
    No & 62\% & 63\% & 62\% & 62\% \\ 
    Yes & 63\% & 63\% & 63\% & 63\%   \\ \hline
    \end{tabular}
\end{table}

Table~\ref{tab:results_proc} summarizes the results, showing that pre-processing slightly improved the model's performance. This improvement can be attributed to the reduced average token length or to the fact that SCoPE removed all function and variable names. This forced the model to learn without relying on these potentially misleading identifiers, which are often defined by programmers and may not reliably reflect the intent of the underlying code. As a result, the model focused on understanding the root cause of vulnerabilities rather than relying on token names, resulting in more robust and generalized learning. 

Building on this, LLaMA 3.2 was further fine-tuned on the processed dataset for four epochs, resulting in performance improvements. These improved results are summarized in Table \ref{tab:results_diversevul}. Although not originally designed for code-specific tasks, LLaMA 3.2 demonstrates competitive performance when fine-tuned for vulnerability detection.

\begin{table}[!htbp]
    \centering
    \caption{Performance of NatGen and LLaMA 3.2.}
    \label{tab:results_diversevul}
    \begin{tabular}{l@{\hskip 0.5cm}c@{\hskip 0.5cm}c@{\hskip 0.5cm}c@{\hskip 0.5cm}c}
    \hline
    \textbf{Model} & \textbf{Accuracy} & \textbf{Precision} & \textbf{Recall} & \textbf{F1-Score} \\ \hline
    NatGen \cite{Chen2023} & 92\% & 52\% & 43\% & 47\% \\
    LLaMA 3.2 & 66\% & 65\% & 67\% & 66\% \\ \hline
    \end{tabular}
\end{table}

We compared our results with the best AI model reported by the authors of the DiverseVul dataset, who used NatGen for the same purpose as our model, SVD. While the differences in models, datasets, and pre-processing prevent a direct comparison, this provides a baseline for performance. LLaMA achieves lower accuracy, however, it outperforms NatGen in terms of precision, recall, and F1-Score. Since accuracy is heavily influenced by the number of classes tested and the imbalance of the original dataset, as shown in Fig.~\ref{fig:data_distr}, it cannot be used as a reliable metric without proper balancing. The authors of the DiverseVul dataset noted that errors may have occurred during the merging of multiple datasets, which might have introduced label noise. In our study, we used SCoPE to remove erroneous data points from the DiverseVul dataset, which may explain the improved model performance.

\section{Threats to Validity}
\label{sec:threats}
The threats to the results' validity can essentially be found in the data utilized to train the models. Since the process of identifying erroneous entries is automated, some unidentified faulty data points may still exist in the dataset. Additionally, this study did not consider if the selected LLM was pre-trained on data contained in the dataset utilized for fine-tuning. If this is the case, it could hurt performance.  

\section{Conclusions and Future Work}
\label{sec:conclusions}

This study explores the usage of DiverseVul in the context of LLMs for SVD. The dataset was curated using the SCoPE framework, which played a critical role in identifying and eliminating erroneous data points. Those points were removed, and a refined version of the dataset was built and will be made publicly available. To the best of the authors' knowledge, the erroneous data identified in DiverseVul has not been reported in previous research, making this study a valuable contribution to the field.  

While evaluating the performance of LLaMA 3.2 on SVD, the impact of various pre-processing techniques was explored. Fine-tuning the model on the processed dataset demonstrated an improvement in detection performance. This pre-processing step allowed the model to focus more effectively on the semantic structure of the code, improving its ability to identify vulnerabilities and minimizing distractions caused by irrelevant elements. This success motivated further experimentation, which included increasing the number of training epochs, as the initial training approach did not fully stabilize the model's loss. This adjustment led to a 3\% improvement, leading to a 66\% of F1-Score, demonstrating that LLaMA 3.2 can achieve competitive performance in SVD tasks.

Several promising directions for future work can be explored. One direction is to expand the scope of the study by using a larger portion of the DiverseVul dataset, as only a subset was used in this initial research. Extending the training time and increasing the number of epochs could further optimize the model's performance, as resource constraints limited the potential for refinement in this study. Another direction could be to explore the robustness of the model, as the results suggest that programmer-controlled elements may influence its predictions.

Given recent advances in large language models, it would also be valuable to compare the performance of other state-of-the-art models, such as DeepSeek-r1, on SVD task. Such comparisons could provide deeper insights into their relative strengths and weaknesses. In addition, advanced techniques such as prompt engineering and retrieval-augmented generation could be explored to further improve both the performance and explainability of the model when applied to SVD tasks.

\begin{credits}
    \subsubsection{\ackname} This work was done and funded in the scope of the BEHAVIOR project  (NORTE2030-FEDER-00576300 no. 14391). This work was also supported by UIDB/00760/2020.

    \subsubsection{\discintname}
    The authors have no competing interests to declare that are  relevant to the content of this article. 
\end{credits}
%
%
%
\bibliographystyle{splncs04}
\bibliography{bibliography}
\end{document}